\documentclass[sigconf]{acmart}

\usepackage{graphicx}
\usepackage{subfigure}
\usepackage{booktabs} 
\usepackage{mathtools}
\usepackage{comment}
\usepackage{multirow}
\usepackage{amsmath,amsfonts}
\usepackage{algorithmic}
\usepackage{textcomp}
\usepackage{xcolor}
\usepackage{xspace}
\usepackage{fancyhdr}
\usepackage{makecell}

\usepackage{hyperref}

\newcommand{\PCignore}[1]{}

\def\Snospace~{\S{}}

\newcommand{\squishlist}{
 \begin{list}{$\bullet$}
  { \setlength{\itemsep}{0pt}
     \setlength{\parsep}{3pt}
     \setlength{\topsep}{3pt}
     \setlength{\partopsep}{0pt}
     \setlength{\leftmargin}{1.5em}
     \setlength{\labelwidth}{1em}
     \setlength{\labelsep}{0.5em} } }

\newcommand{\squishlisttwo}{
 \begin{list}{$\bullet$}
  { \setlength{\itemsep}{0pt}
     \setlength{\parsep}{0pt}
    \setlength{\topsep}{0pt}
    \setlength{\partopsep}{0pt}
    \setlength{\leftmargin}{2em}
    \setlength{\labelwidth}{1.5em}
    \setlength{\labelsep}{0.5em} } }

\newcommand{\squishend}{
  \end{list}  }

\definecolor{amber}{rgb}{0.75,0.35,0.0}

\newcommand{\system}{DNNFuser\xspace}
\newcommand{\newgamma}{G-Sampler\xspace}

\AtBeginDocument{%
  \providecommand\BibTeX{{%
    \normalfont B\kern-0.5em{\scshape i\kern-0.25em b}\kern-0.8em\TeX}}}

\acmConference[Conference acronym 'XX]{Make sure to enter the correct
  conference title from your rights confirmation emai}{June 03--05,
  2018}{Woodstock, NY}
\acmBooktitle{ }
\acmPrice{ }
\acmISBN{ }

\begin{document}

\title{\system: Transformer as a Generalized Mapper for Fusion in DNN Accelerators}

\author{Sheng-Chun Kao}
\authornote{Both authors contributed equally to this research.}
\affiliation{%
  \institution{Georgia Institute of Technology}
  \city{Atlanta}
  \country{GA}}
\email{felix@gatech.edu}

\author{Xiaoyu Huang}
\authornotemark[1]
\affiliation{%
  \institution{Georgia Institute of Technology}
  \city{Atlanta}
  \country{GA}}
\email{xhuang394@gatech.edut}

\author{Tushar Krishna}
\affiliation{%
 \institution{Georgia Institute of Technology}
  \city{Atlanta}
  \country{GA}}
\email{tushar@ece.gatech.edu}


\begin{abstract}
Dataflow/mapping decides the compute and energy efficiency of DNN accelerators. Many mappers have been proposed to tackle the intra-layer map-space. However, mappers for inter-layer map-space (aka layer-fusion map-space), have been rarely discussed. 
In this work, we propose a mapper, \system, specifically focusing on this layer-fusion map-space. While existing SOTA DNN mapping explorations rely on search-based mappers, this is the first work, to the best of our knowledge, to propose a one-shot inference-based mapper. We leverage Transformer as our DNN architecture to learn layer-fusion optimization as a sequence modeling problem. Further, the trained \system can generalize its knowledge and infer new solutions for unseen conditions. Within one inference pass, \system can infer solutions with compatible performance to the ones found by a highly optimized search-based mapper while being 66x-127x faster.

\end{abstract}



\settopmatter{printacmref=false} 
\renewcommand\footnotetextcopyrightpermission[1]{} 
\pagestyle{plain} 

\maketitle

\section{Introduction}
Accelerators for Deep Neural Network (DNN) models are 
commonplace today across the system stacks from cloud clusters~\cite{tpu_spec} to edge devices~\cite{edge_tpu,eyeriss_isca,nvdla}. For computation and energy efficiency, different dataflows/mappings~\cite{nvdla, eyeriss_isca} have been proposed to optimize the data movement and compute utilization inside DNN accelerators. Many mappers have been proposed to automate this mapping optimization problem~\cite{gamma, mindmapping,cosa}.

The conventional map-space of SOTA DNN mappers follows the assumptions of layer-by-layer execution in DNN accelerators~\cite{eyeriss_isca,gamma,mindmapping}. Each layer of DNNs is mapped onto the accelerator sequentially and iteratively. The output activations are streamed out to the off-chip memory. Upon completion, those activations as inputs of the next layer are streamed back to on-chip buffer and computation units. This scheme often introduces a large number of off-chip memory accesses.
With the increasing interest in high-resolution image processing or long-sequence language models, these off-chip access overheads start to stick out. Some prior researchers~\cite{alwani2016fused, wei2018tgpa, flat} have shown that a rarely explored mapping dimension could potentially ameliorate this off-chip access challenge ---inter-layer dataflow (or so-called layer fusion).

Instead of sticking to layer-by-layer execution, layer fusion opens the discussion of mapping multiple layers to the DNN accelerators simultaneously to leverage the immediate data reuse of intermediate activations. Fused-layer CNN~\cite{alwani2016fused}, TGPA~\cite{wei2018tgpa}, and FLAT~\cite{flat} showcased huge potential benefits, however relying on their manual-designed layer fusion strategies. While many mappers are proposed to automate the search in intra-layer map-space, automated mappers for the layer fusion map-space have been rarely discussed.

In this paper, we propose \system, which is a Transformer-based mapper targeting the map-space of DNN layer-fusion. \system is an orthogonal work to existing intra-layer mappers~\cite{gamma, mindmapping,cosa}. We envision the proposed framework being further combined with the existing SOTA intra-layer mappers (as part of future work) to harvest and jointly explore both axes of performance improvement.
We summarize the key technical innovations within \system as follows:

\textbf{i) Layer-fusion Map space.}  \system is one of the first automated mappers to target the layer-fusion map-space while most prior work focuses on intra-layer map-spaces~\cite{gamma, mindmapping,cosa}.

\textbf{ii) Transformer-based Mapper.} To the best of our knowledge, \system is the first mapper for DNN accelerators leveraging Transformer pre-training. The key feature is: once the mapper is trained, it could produce a mapping strategy at inference-time, i.e., no additional search process is needed. Many existing DNN mappers rely on optimization methods such as genetic algorithm (GA)~\cite{gamma}, Bayesian Optimization (BO)~\cite{hasco}, Mixed-Integer-Programming (MIP)~\cite{cosa}, or RLs~\cite{flextensor}, which work well but require large search time. A recent work~\cite{mindmapping} shows gradient-based surrogate model could reduce the search time, however, a search process is still needed. In this work, we model the layer-fusion mapping optimization as a sequence modeling problem and leverage the Transformer as our underlying DNN architecture. 

\textbf{iii) Generalizability.} We show that \system can generalize its knowledge to unseen HW conditions. In our study, we train \system to produce an optimized mapping that can condition on the current available on-chip buffer in the DNN accelerator. Note that the available on-chip buffer could vary since other small kernels could be running concurrently and occupying some portion of the on-chip buffer. Whenever the HW constraint (on-chip buffer) changes, the mapping needs to be updated, which means another search process for the search-based methods. In contrast, for \system, we can produce a new mapping at inference time.

\textbf{iv) Transfer Learning.} 
\system can be trained on several DNN workloads and serve as a general mapper. We show that the trained general mapper can execute transfer learning on a new DNN workload with only a few epochs of fine-tuning, demonstrating a strong sign of transferable generalized knowledge.



\section{Background on DNN Accelerators}

DNN accelerator design points often can be described with two parts: HW resources and the mapping strategy

\begin{figure}
\begin{center}
\includegraphics[width=1\linewidth]{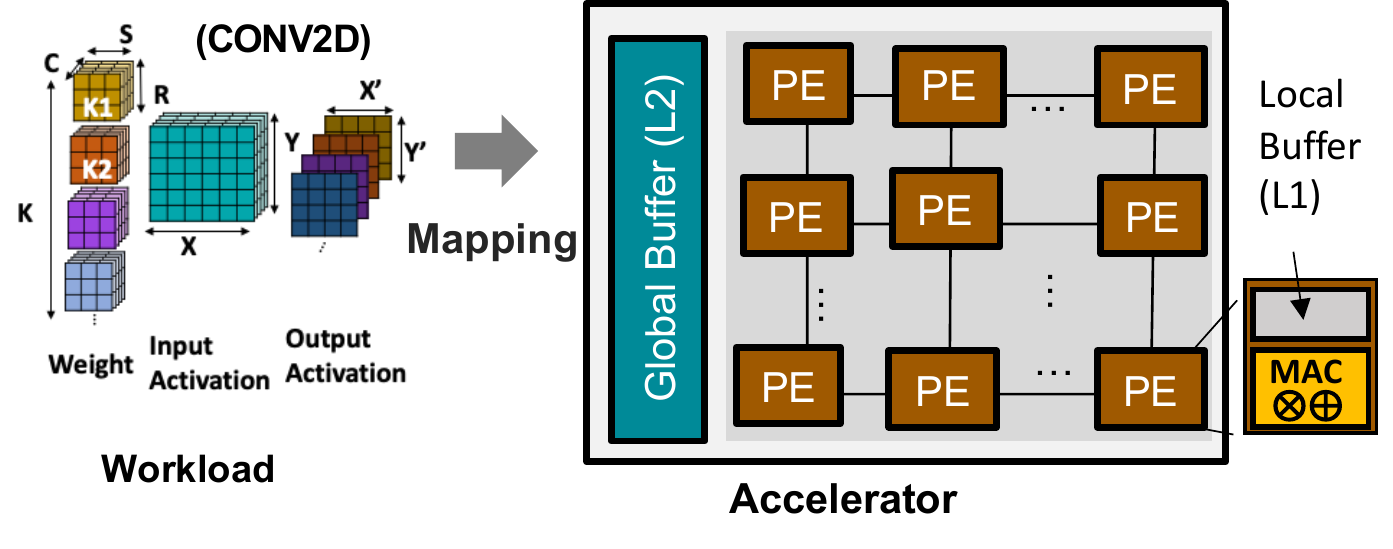}
\end{center}

\caption{A typical layer-by-layer execution scheme on a canonical DNN accelerator.}

\label{fig:accelerator}
\end{figure}

\textbf{Hardware Resources.}
Spatial DNN accelerators~\cite{eyeriss_isca} comprise an array of processing elements (PEs) (\autoref{fig:accelerator}). Each PE has a MAC to compute partial sums and a scratchpad to store weights, activations, and partial sums. The accelerators also house a shared on-chip (global) buffer to pre-fetch activations and weights from off-chip memory for the next tile of computation that will be mapped over the PEs.
Networks-on-Chip (NoCs) are used to distribute data from the on-chip buffer to PEs, collect the partial or full outputs, and write them back to the on-chip buffer.

\textbf{Intra-layer Mapping.}
Conventionally, we use mapping/dataflow to refer to intra-layer mapping~\cite{eyeriss_isca, gamma}, which includes (1) tiling (how tensors are sliced, stored, and fetched across the memory hierarchy), (2) ordering (the order in which loop computations are performed), (3) parallelism (how compute is mapped across PEs in space), and (4) clustering (how to compute/buffer are structured into a hierarchy of levels). These form a search space as large as $O(10^{24})$~\cite{gamma}, which is not feasible for exhaustive search. Hence there is a wide array of works on searching through this map-space~\cite{gamma,mindmapping,cosa,flextensor}.

\textbf{Inter-layer Mapping: Layer Fusion.}
Inter-layer mapping (layer fusion\footnote{Layer fusion means we fuse the computation of two or more layers/operators together and map them onto the accelerator simultaneously so that their activations are staged on-chip and reused immediately.}) is an orthogonal map-space to intra-layer mapping. It decides a special tiling dimension across multiple dependent DNN layers. The layer fusion map-space grows exponentially with the number of layers in a DNN workload. For example, if we allow 64 tiling choices per layer (which we use in our evaluations), for an 18 layers DNNs such as Resnet18~\cite{Resnet}, the map-space will become as large as $64^{18}$=$O(10^{32})$. With deeper DNN workloads such as Resnet50, Mobilenet-V2, and Mnasnet, the map-space grows to the order of $O(10^{90})$. In this map-space, prior works introduced manually-tuned fusion strategies~\cite{alwani2016fused, wei2018tgpa, flat}, and no prior mappers, to the best of our knowledge, search this space automatically.



\section{Problem Formulation}
\label{sec:problem_formulation}
In the DNN accelerator, the on-chip buffer is often limited and not able to stage the full output activation. The key idea of layer fusion is that, rather than streaming back the full output activation to the off-chip memory like in ~\cite{gamma, mindmapping, cosa}, we stage ``partia'' output activation on-chip to exploit data locality and thus boost data reuse. 

Practitioners start to use ``micro-batching'' as the go-to strategy. With micro-batching, we will divide a batch of activations into multiple micro-batches. In practice, We search for the largest acceptable micro-batch size that allows us to stage all intermediate activations on-chip, which becomes the most naive micro-batching strategy. 
However, this naive strategy cannot maximize the reuse opportunity. Different layers produce different sizes of output activations, and a naive unified micro-batch size could leave the on-chip buffer under-utilized.

\emph{A more sophisticated layer-wise micro-batching strategy is needed to maximize the on-chip buffer utilization, minimize off-chip access, and finally improve the runtime (latency or throughput) performance.}



\begin{figure*}
\begin{center}
\includegraphics[width=1\linewidth]{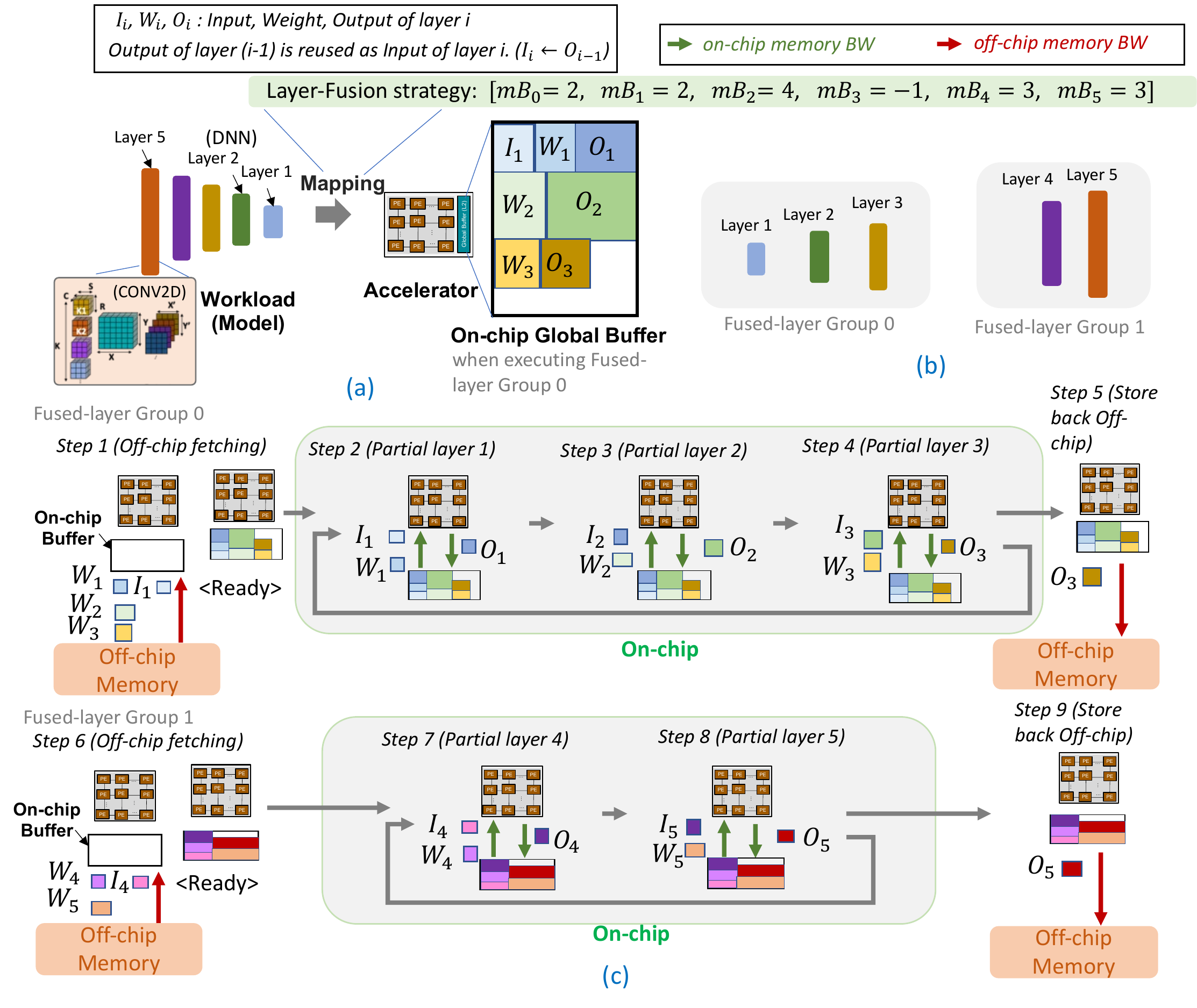}
\end{center}

\caption{A layer-fusion mapping example on a 5-layer DNN model. (a) The hardware implication, (b) the introduced two group of fused-layer, and (c) the execution steps of each group.}

\label{fig:fusion_example}
\end{figure*}

We call a layer-wise micro-batching strategy --- \textit{layer fusion strategy}, since it is essentially dividing DNN layers into multiple fused-layers. For example, in \autoref{fig:fusion_example} we have a layer fusion strategy for a 5-layer DNN workload. The strategy dictates the output micro-batch sizes of each layer, where the first value represents the input micro-batch size. We use "-1" to represent a signal to synchronize and stream data back to off-chip memory before proceeding to the next layer. This synchronization divides the dataflow into two groups of fused-layers, as shown in \autoref{fig:fusion_example}(b). The groups of fused-layers will be executed sequentially, as shown in \autoref{fig:fusion_example}(c). \autoref{fig:fusion_example}(a) shows a snapshot when the accelerator is executing fused-layer group-0. For an N layer DNN workload, the layer fusion strategy is represented as follows
\begin{align}
\hskip\parindent & \begin{gathered}
    Strategy=[mB_{0}, mB_{1},...mB_{N}]
    \end{gathered} 
\label{formula:strategy}
\end{align}
, where $mB_{i}$ is the micro-batch size of layer i. To search for the optimum the micro-batching (layer-fusion) strategy, the problem formulation is as follows:

\textbf{\textit{Problem formulation: Given a DNN workload, batch size, and available on-chip memory, produce a layer-fusion strategy that optimizes the end-to-end latency or throughput performance.}}

\section{\system}
\begin{figure*}
\begin{center}
\includegraphics[width=1\linewidth]{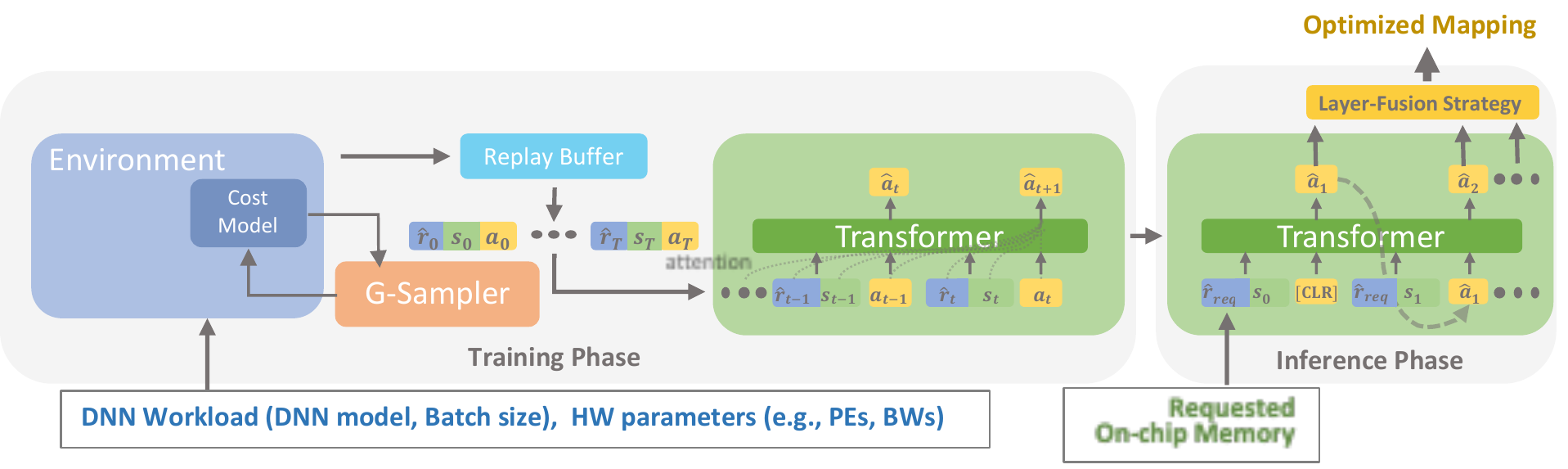}
\end{center}

\caption{The workflow of \system.}

\label{fig:algorithm}
\end{figure*}

We propose \system, a pre-trained Transformer-based mapper for layer fusion optimization for DNN workloads. \system, with a fully trained model, features the ability to infer an optimized mapping for different HW conditions at inference time. As shown in \autoref{fig:algorithm}, \system takes in the input of a workload (a DNN model and its batch size), HW parameters (number of PEs, on-chip BW, off-chip BW), and an HW condition (requesting on-chip buffer usage), and outputs an optimized mapping.

\subsection{Motivation for using RLs and Transformers}
 
The key objective is to train a generative model (a decoder model, or a encoder-decoder model where encoder can take in the condition) so that it can serve as a mapper while not incurring any search process. DNNs have been proven repeatedly to have some amount of ``knowledge generalization'' ability. Deep RLs (DRLs) become one of the best testimonies of this ``generalizability''. Even though the number of possible frames (or states) in RL environments, such as Atari games and OpenAI gyms, is often too large to be fully enumerated (similar challenge as in our layer fusion map-space), a well-trained DRL can play well in those environments and generalize its knowledge to states never seen before. This motivates us to take advantage of DRL in our layer fusion problem. 

However, DRLs are also known for their ``unstableness'', as they rely on NNs to learn the ``connection'' of each state itself in a long trajectory while also suffering from the delayed reward and vanishing gradient problems 
(which is even more challenging in our layer fusion problem (discussed in detail in \autoref{sec:why_teacher_model})), and are also sensitive to the case-by-case setting of discount rate~\cite{decision_transformer}. Transformers, meanwhile, are known for their powerful ability to make customized ``connections'' by attention mechanisms~\cite{vaswani2017attention} even in long-sequence scenarios. 
Recent research~\cite{decision_transformer,boustati2021transfer,ortega2021shaking} shows that learning RL problems with a Transformer is promising and ``stable'' because it can automatically learn how to attend to previous states and also avoid the discount rate problem.

\emph{Therefore, we decide to formulate layer fusion into an RL problem for the potential ``generalizability'' and utilize Transformers to learn the formulated RL problem for its ``stability''. }

\subsection{Formulating Layer Fusion into RL Problem}
\label{sec:rl-formulation}


We model a micro-batching decision for a layer in the DNN workload as one time-step in the RL environment. 
For an N layers DNN, the agent (\system) would go through N+1 time-steps, visiting N+1 states, making N+1 actions, and receiving N+1 rewards, before it receives a ``done'' signal and complete one trajectory. 

\textbf{Action.} The action at time-step t is the micro-batch size of layer-t ($mb_{t}$), also note as $a_{t}$(=$mb_{t}$).

\textbf{State.} We formulate a state at time-step t as follow,
\begin{align}
\hskip\parindent & \begin{gathered}
    s_{t}=[K_{t}, C_{t}, Y_{t}, X_{t}, R_{t}, S_{t}, \widehat{M}, P_{a_{0}..a_{t-1}}]
    \end{gathered} 
\label{formula:state}
\end{align}
The first 6 dimensions are the tensor shape of the current DNN layer. We use the 6-loop notation of CONV layer (other types of layers can also be mapped and represented with these 6-loop notations). $K_{t}$, $C_{t}$ are output and input channels. $Y_{t}$, $X_{t}$ are height and width of activations. $R_{t}$, $S_{t}$ are height and width of weight kernels. $\widehat{M}$ is given as the currently available memory (normalized by the batch size). $P_{a_{0}..a_{t-1}}$ designates the runtime performance (latency or throughput) after a series of actions: [$a_{0}$, ..., $a_{t-1}$]. For example at time-step-0, where no previous actions have been made, $P_{0}$ will be the runtime performance before any layer fusion strategy is performed. At time-step-1, $P_{a_{0}}$ is the runtime performance when we only apply layer fusion strategy to the first layer, and so on.


\subsection{Transformers for RL-based Layer-Fusion Problem}
Reinforcement learning is often noted as harder to converge, higher variance in the result, and unstable in training. Thus, there remains a long-lasting hope of converting RL into a supervised learning problem. Recently, an array of works~\cite{decision_transformer,boustati2021transfer,ortega2021shaking} leverage Transformers to model off-line reinforcement learning and achieve SOTA performance in many typical RL tasks. Leveraging similar taxonomy, we convert our RL-based layer fusion formulation into a Transformer-based supervised learning problem as follows.

\subsubsection{Formulation}
We convert RL state transition into a sequence like a sentence in language models (similar to ~\cite{decision_transformer,boustati2021transfer,ortega2021shaking}). One trajectory of state transition is a sequence of reward, state, action pair, ([($\widehat{r_{0}}$,$s_{0}$,$a_{0}$), ..., ($\widehat{r_{t}}$,$s_{t}$,$a_{t}$), .., ($\widehat{r_{N}}$,$s_{N}$,$a_{N}$)), with N being the number of layers of the targeting DNN models.
The Transformer will take this sequence as input and generate a prediction for the next action, $\widehat{a}$, as shown in \autoref{fig:algorithm}. For supervised learning, the loss is taken as the Mean Square Error between the predicted action, $\widehat{a_t}$, and the actual action, $a_t$, for $t\in \mathopen[0, 1, ..., N\mathclose]$.

\subsubsection{Discussion}
There is two high-level intuition for why this formulation is specifically useful in our problem.
\squishlist
\item 1) Sparse and distant reward. The correlation between action and reward is often sparse and distant in the layer fusion scenario. The optimal micro-batch and the corresponding impact on the overall memory usage for the current layer often do not depend closely on the last action (last micro-batch decision), but those several time-steps back. The transformer is expected to efficiently deal with the long-sequence scenario and excavate the relation between each action pair, near and far. 
\item 2) The attention layer in the Transformer is capable of finding shortcuts between individual trajectories~\cite{decision_transformer}. This means that our \system can find paths that are missing from the dataset, and more specific to our layer-fusion scenario, some peak memory usages that it has never seen. 
\squishend


\subsubsection{Conditional Reward}

Rather than the common practice of taking only state or state-action pairs to generate a policy in traditional RLs~\cite{a2c,ddpg}, we take the full reward-state-action pair as inputs as previously described. The benefit is that the reward is now exposed as input at the inference phase \autoref{fig:algorithm}), which opens the opportunity to control the output (mapping solution) by the ``desired'' reward. This technique is also leveraged in ~\cite{decision_transformer}.

We use ``requesting/conditioning on-chip memory usage'' as the conditioning (desired) reward (noted as $\widehat{r_{t}}$). This formulation brings three benefits:
\squishlist
\item 1) The current available on-chip memory sizes are known parameters, therefore we could use it as a reasonable condition.
\item 2) The heuristics tell that a layer fusion strategy that maximizes the on-chip memory usage often achieves better runtime performance. Therefore, we can expect a high-performance solution if conditioning on the total available memory. 
\item 3) By setting memory usage as a condition, it also set up a potential use case of generalizing to unseen memory sizes, discussed later in \autoref{sec:generalization}.
\squishend

\subsection{Training Data Collections}
\label{sec:pre-training}


\subsubsection{Imitation Learning and Teacher Model}
\label{sec:why_teacher_model}
An DNN learning trick, ``imitation learning'', is found useful in training Transformers~\cite{decision_transformer,boustati2021transfer,ortega2021shaking}. ``Imitation learning'' could be imitating and learning 1) from self previous experiences as used in related works \cite{decision_transformer,boustati2021transfer,ortega2021shaking} or 2) from another well-trained teacher model\footnote{Teach model will teach the student model (model in training) to learn faster by demonstration}, which we use in this work and find it critical for Transformers to learn well in our layer fusion problem.


In previous work, such as Decision Transformer (DT)~\cite{decision_transformer}, the ``imitation learning'' dataset is collected by an RL agent sampling in the targeting Atari or OpenAI environment. However, we find that popular RL algorithms, including Advantage Actor-Critic (A2C)~\cite{a2c} and Deep Deterministic Policy Gradient (DDPG)~\cite{ddpg}, suffer from inefficient sampling when interacting with the layer fusion map-space, and their trajectories are therefore inefficient demonstrations for ``imitation learning''. 

We observe that, unlike Atari or OpenAI with smooth state transitions, in our mapping problem, state transits abruptly between two consecutive time-steps (e.g., the state includes information of current layer shape, which does not have deducible relation to the layer shape in next state), and that the variance is detrimental to the convergence of the mentioned RL algorithms. 
Thus, rather than letting the agent (our model) explore in the environment itself to collect ``experiences'', which is particularly inefficient in this case,
we decide to use a teacher model, with some prior knowledge, to demonstrate directly some good trajectories. 
Next, we will discuss the details of our teacher model.

\subsubsection{Developed Teacher Model: \newgamma}
~\label{sec:gamma_teacher}
Gamma~\cite{gamma} is one of the SOTA intra-layer mappers for DNN accelerators. We extended Gamma to support inter-layer layer-fusion map-space and found it several orders of magnitude better in performance than other optimization methods such as A2C~\cite{a2c}, PSO~\cite{pso_paper}, DE~\cite{de}, and so on~\cite{cma, tbpsa, bayes, ddpg, reinforce}, which we show in our evaluations (\autoref{sec:comarisons}). We use the developed extended Gamma as our teach model and call it ---\newgamma. We use \newgamma to search for optimal layer-fusion strategy given different on-chip buffer sizes. 


Note that we empirically found that \newgamma works well as a layer fusion mapper, which beats many optimization methods. However, \newgamma is still a search-based method, which is inevitably slower than most of the inference-based DNN models.
Therefore, \newgamma only serves as a training data generator, which samples several good solutions (trajectories) in the layer-fusion environment (\autoref{fig:algorithm}). \system, as a student, learns from these demonstrations and generalizes the knowledge to learn to generate optimized mapping at inference time. 


\subsection{Model Training and Inference}
\subsubsection{Training}
With all the above-mentioned preparation and setup, we can now train our model. As shown in \autoref{fig:algorithm}, the steps are as follows. 1) Data collection with teacher model. We take \newgamma and ask it to search for several (4-10) sets of optimized mapping in different conditions (on-chip memory sizes).
2) Formulating into RL state transition. We take those solutions, which is a sequence of actions in RL terminology, and decorate them with the state and reward information to make it a state transition trajectory. These decorated trajectories will be stored in the replay buffer (essentially a place to house the training dataset).
3) Model training. We sample the training data from the replay buffer and train our model in an "imitation learning" fashion. 

\subsubsection{Inference}
At inference time, the model takes in a conditioning reward, which is the conditioning on-chip buffer usage, and generates a sequence of actions as a solution in an auto-regressive manner. The actual on-chip buffer usage of the solution adheres to the desired condition. 

\subsection{Generalization and Transfer Learning}
\label{sec:generalization}
\subsubsection{Generalization} After training, \system can work on not only the seen conditions (on-chip buffer sizes) but also some unseen conditions. For example, we can train \system with some conditioning on-chip buffer sizes, such as 16, 32, 48, and 64 MB, and it could generalize its knowledge to the unseen interpolated buffer sizes between 16-64MB. For example, if the available buffer sizes in the HW suddenly change from 32MB to 28MB because some amount of buffer is occupied by other small procedures. We do not need to re-train the model (or launch a new search as we will need in a traditional search-based method). We can simply do another inference with 28MB as the new condition, and \system with its generalized knowledge could infer a solution.

\subsubsection{Transfer Learning} \system can be trained on one or several common DNN workloads and serve as a pre-trained model for transfer learning. For a new DNN workload, we can take this general model and fine-tune it on the new workload. The fine-tuning process only takes a few epochs to converge since the generalized knowledge in \system can be transferred and utilized in the new workload. Empirically, we find that fine-tuning process requires only 10\% of the training epoch compared to training from scratches, and can achieve compatible or better performance than a training-from-scratch model.

\label{sec:finetuning}

\section{Evaluation}

\subsection{Setup} 
\label{sec:setup}
\textbf{Workload.} We experiment on several common DNN workloads, including VGG16~\cite{vgg}, Resnet18~\cite{Resnet}, Resnet50~\cite{Resnet}, Mnasnet~\cite{mnasnet}, and Mobilenet-V2~\cite{sandler2018mobilenetv2}.

\textbf{Accelerator Assumption.} We experiment on the DNN accelerator configurations of 1024 PEs, on-chip buffer sizes of 64 MB, off-chip memory BW of 900GB/s, on-chip memory BW of 9000GB/s, and frequency of 1GHz, which is similar to ~\cite{tpu_spec, eyeriss_isca}. We set this as our underlying DNN accelerator modeled by our cost model. 

\textbf{Cost Model.}
We built an analytical cost model to model the effect of layer fusion in DNN accelerators. The cost model focuses on modeling interactions between layers and assumes the ideal performance for intra-layer map-space, which can already be achieved by existing intra-layer mappers~\cite{mindmapping,gamma}.   
The cost model takes in a DNN workload, the number of PEs, on-chip memory size, on-chip and off-chip memory bandwidths (BW), and a layer fusion strategy, and returns its runtime performance and peak memory usage. The built cost model is validated against MAESTRO~\cite{maestro}, which is in turn validated again real chip performance.

\textbf{Baseline Mapping.} Our baseline mapping, no-fusion mapping, is the best possible mapping without exploring layer fusion map-space (i.e., the most ideal mapping in intra-layer map-space), representing mapping that can be found by existing SOTA mappers~\cite{gamma, mindmapping, cosa}. To compare the effectiveness of the different searching algorithms for this fusion strategy search problem, we compute the speedup of the solution proposed by each searching algorithm (a fusion-based mapping) to the baseline mapping (no-fusion mapping).


\textbf{Baseline Search Methods.} 
Among optimization methods, we compare with many widely used ones as our baseline search methods for the inter-layer mapping, including Particle Swarm Optimization (PSO), Covariance matrix adaptation evolution strategy (CMA-ES), Differential Evolution (DE), Test-based population-size adaptation (TBPSA) and standard GA (stdGA), whose implementations are from nevergrad~\cite{nevergrad} by Facebook. We also implement an RL method A2C~\cite{a2c}. We allow all of them to have sampling budgets of 2K.

\textbf{Baseline Sequence Model.}
\system is a Transformer-based sequence models. To show the effectiveness of Transformer as the sequence model for this problem, We also implement an RNN-based sequence-to-sequence model (Seq2Seq) as a baseline sequence model. The Seq2Seq is made of a LSTM with 2 layers of fully connected layers and 128 hidden dimension in each encoder and decoder.

\textbf{Performance Metric.} To fairly compare the effectiveness of optimization methods, we compare the achieved speedup of their layer-fusion solutions over the baseline mapping.

\textbf{Implementation Details of \newgamma.} \newgamma, our developed search method (\autoref{sec:gamma_teacher}), has a population size of 40 and runs for 50 generations, which also ends at a sampling budget of 2K as baselines do. 

\textbf{Implementation Detail of \system.} 
\system is made of three transformer blocks, two heads, and a hidden dimension size of 128. We train the model for 100K epochs.

\begin{table}[t]

  \centering
  \caption{Performance comparisons of different optimization methods on VGG16 workload with two different cases of on-chip memory constraints. The speedup is the latency of the found mapping by each search algorithm to the latency of the baseline mapping (described in \autoref{sec:setup}.) All the experiments run on a 4-core Intel Xeon CPU with a Nvidia GTX 1070 GPU.
}
  \vspace{-0.3cm}
    \begin{tabular}{c|c|c|c}\hline
    Algorithm & Speedup & Act. Usage (MB) & Search Time (mins) \\\hline
   \textit{Case-1}  & \multicolumn{3}{c}{On-chip Memory constraint 20M, Batch size 64} \\\hline
            PSO   & N/A     & 102.76         & 69.17 \\
            CMA   & N/A     & 186.25          & 77.03 \\
            DE    & N/A     & 114            & 65.17 \\
            TBPSA & N/A     & 153.34          & 110.50 \\
            stdGA & N/A     & 139.69          & 61.66 \\
            A2C   & 0.98    & \textbf{2.26}            & 335.63 \\\hline
     \newgamma    & 1.19  & \textbf{16.46}          & 0.66 \\
     Seq2Seq          & 1.05   &  16.06           & 0.01\\
    \system      & \textbf{1.20}    & \textbf{19.27}          & \textbf{0.01} \\\hline
    \textit{Case-2}  & \multicolumn{3}{c}{On-chip Memory constraint 40M, Batch size 128}\\\hline
    PSO       & N/A            & 255.3       & 93.28 \\
    CMA       & N/A            & 411.04      & 91.42 \\
    DE        & N/A            & 149.32      & 104.74 \\
    TBPSA     & N/A            & 245.66      & 106.20 \\
    stdGA     & N/A            & 236.03      & 151.74 \\
    A2C       & N/A            & 372.51    & 293.81 \\\hline
   \newgamma & 2.06          & \textbf{37.73}   & 1.27 \\
    Seq2Seq  & 1.51          &  35.4           & 0.01\\
   \system   & \textbf{3.13}            & \textbf{37.73}  & \textbf{0.01} \\\hline
    \end{tabular}%

  \label{tab:exp1}%
\end{table}%
\begin{table}[htbp]

  \centering
  \caption{Speedup performance of \system (DF) and Seq2Seq (S2S) on unseen conditioning memory usage (20, 25, 40, 35, 40, and 45MB). \system is only trained on conditioning memory usage of 16, 32, 48, and 64MB.}
  \vspace{-0.3cm}
    \begin{tabular}{c|c|c|c|c|c|c}\hline
    \multicolumn{1}{l}{ \makecell[l]{Model \\ (Batch=64)}} & \multicolumn{3}{|c|}{VGG16} & \multicolumn{3}{c}{Resnet18} \\\hline
     \makecell[l]{Cond. Mem.\\ Usage (MB)} & DF & S2S &  \makecell[c]{G-\\Sampler} & DF & S2S & \makecell[c]{G-\\Sampler} \\\hline
    20    & 1.20 & 1.04  & 1.19  & 1.27 & 1.32 & 1.37 \\
    25    & 1.20 & 1.04  & 2.18  & 1.27 & 1.32 & 1.34 \\
    30    & 1.16 & 1.83  & 1.86  & 2.31 & 1.56 & 1.51 \\
    35    & 1.88 & 1.85  & 2.14  & 2.31 & 1.56 & 1.53 \\
    40    & 1.97 & 1.86  & 2.17  & 2.68 & 1.56 & 2.88 \\
    45    & 1.97 & 2.02  & 2.30  & 2.68 & 1.56 & 2.95 \\\hline
    \end{tabular}%

  \label{tab:exp2}%
\end{table}%


\begin{table}[htbp]

  \centering
  \caption{Speedup comparisons of Transfer-DF (\system pre-trained on VGG16 and Resnet18 and tranfer learning on new workloads), Direct-DF (\system trained from scratch on new workload), and GS (\newgamma running full-search) on different conditioning memory usage.}

    \begin{tabular}{c|c|c|c}\hline
    Cond. Mem. Usage (MB) & Transfer-DF &  \,\,\,\,Direct-DF\,\,\,\,  & \,\,\,\,GS\,\,\,\, \\ \hline
    & \multicolumn{3}{c}{Resnet50, Batch size 64} \\\hline
    25    & 1.31  & 1.17  & 1.41 \\
    35    & 1.78  & 1.78  & 1.94 \\
    45    & 2.01  & 2.03  & 2.13 \\
    55    & 2.55  & 2.03  & 2.26 \\\hline
     & \multicolumn{3}{c}{Mobilenet-V2,  Batch size 64} \\\hline
    25    & 1.83  & 1.68  & 2.27 \\
    35    & 2.01  & 1.67  & 2.18 \\
    45    & 2.66  & 2.90  & 2.41 \\
    55    & 2.94  & N/A   & 4.32 \\\hline
     & \multicolumn{3}{c}{Mnasnet,  Batch size 64} \\\hline
    25    & 3.34  & N/A   & 3.60 \\
    35    & 3.34  & 3.34  & 3.17 \\
    45    & 3.34  & 3.34  & 3.82 \\
    55    & 3.46  & 3.53  & 4.07 \\\hline
    \end{tabular}%
  \label{tab:exp3}%
\end{table}%

\subsection{Comparisons with Baselines}
\label{sec:comarisons}
First of all, the sampling efficiency\footnote{The performance improvement over number of samples.} of the optimization algorithms will decide the quality of the found solutions. 
In this experiment, the constraint is the on-chip memory usage, and the objective is to minimize the latency. Many baseline methods such as PSO, CMA, DE, and so on, cannot meet the on-chip memory usage constraint. Note that their memory usage is larger than 20MB (or 40MB at the bottom half the table) in \autoref{tab:exp1}. Therefore their solution is invalid, and we mark the invalid solution as N/A. They cannot find valid solution within the given 2K sampling budget. Given ``enough'' sampling budget, they will eventually learn the constraint and start to optimize the latency. However, 2K sampling budget already take them 1-2 hours. The need of more sampling budget and more hours make these baseline methods not ideal for this problem.  

A2C can successfully find a valid solution within the sampling budget. However, it takes around 5 hours, and its solution ends up worse than the baseline mapping, i.e., speedup smaller than 1.0. 

Next we discuss the choice of sequence model in our framework, it could be the Transformer-based \system or an RNN-based Seq2Seq model. For both sequence models, we leverage the \newgamma to generate the examples (training data for the sequence models). \newgamma itself can converge within around 1.3 mins and achieve 1.19x and 2.06x speedup over the baseline mapping. As for the sequence model, we could observe that both \system and Seq2Seq can perform well in our framework in terms of search time and achieved speedup. However, \system can constantly find better solution, achieving more speedup than Seq2Seq.
\system takes around 0.6-1.3 mins for searching and achieves 1.19x and 2.06x speedup over the baseline mapping. \system can find the solution with compatible or better performance than \newgamma and is \textbf{66x-127x} faster than it.


\subsection{Generalizing to Unseen HW Condition}
We train \system on VGG16 and Resnet18 respectively, using the memory condition of on-chip memory usages of 16, 32, 48, and 64MB. Without re-training, \system can directly generalize its knowledge to unseen HW conditions and infer corresponding solutions, the memory condition of 20, 25, 30, 35, 40, and 45, as shown in \autoref{tab:exp2}. Note that all of these conditions are unseen for the trained \system, and they are the interpolation of the range of memory condition (16 - 64MB) that \system is trained on\footnote{We leave the extrapolation as future work.}. With only one inference, \system and Seq2Seq can both find solutions with compatible performance to \newgamma while \newgamma is running full-search. Among \system and Seq2Seq, we can observe their performance is compatible on VGG16. However, on Resnet18 \system is a clear win. It is because Resnet18 with more number of layers than VGG16 forms a longer sequence, and \system (a Transformer-based model) performs better than Seq2Seq (a RNN-based model) at longer sequence tasks.

\subsection{Transfer Learning}
To demonstrate the ability to execute transfer learning, we train \system on both VGG16 and Resnet18 to form a pre-trained general model. Then we use this general model to execute transfer learning on new DNN workloads: Resnet50, Mobilenet-V2, and Mnasnet. We fine-tune on new workloads (Transfer-DFs) with only 10\% of the training epochs compared to training-from-scratch (Direct-DFs).
As shown in \autoref{tab:exp3}, with only 10\% of training epochs, Transfer-DFs can achieve compatible performance with \newgamma and better performance than Direct-DFs (looking at some edge cases where Direct-DFs fail (N/A)). This tells that prior knowledge gained from pre-training actually increases the training speed and quality, which is especially important when training large DNN workloads such as Resnet50, Mobilenet-V2, or Mnasnet, which would otherwise require more training data.
I.e., Direct-DFs need more teacher demonstrations from the Teacher model than Tranfer-DFs.
This observation showcases the usefulness of transfer learning---with a pre-trained general model of \system, we can learn new workloads faster and better. 

\begin{figure}
\begin{center}
\includegraphics[width=0.9\linewidth]{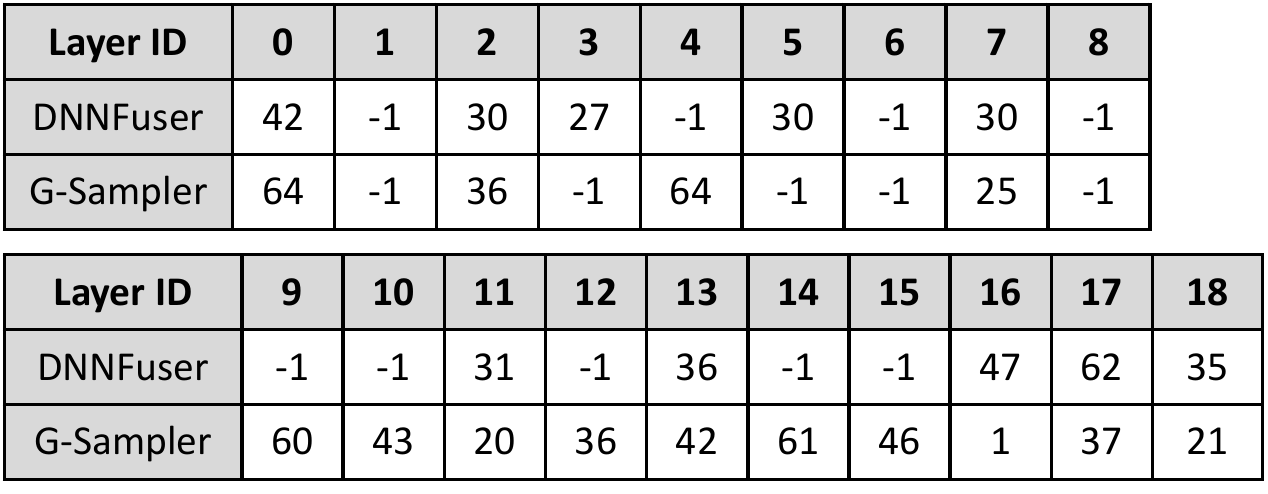}
\end{center}

\caption{The layer-fusion mapping found by \system and \newgamma on ResNet18 with batch size 64 conditioning on memory size of 20MB. Values represent the micro-batch size of each layer, with LayerID 0 being the input to LayerID 1. }

\label{fig:example_sol}
\end{figure}

\subsection{Analysis of Found Solutions}
In \autoref{fig:example_sol}, we show one of the solutions found by \system and \newgamma. There are two main observations: 1) at deeper layers, they learn to fuse more number of layers, since deeper layers tend to have smaller activations. 2) Sudden expansion in either channel depth (e.g. layer 6) or activation shape (due to residual connections) (e.g. layer 8) propels them to synchronize back off-chip because staging on-chip puts too much pressure on the memory capacity.

\section{Related Work}

\textbf{DNN Mapper.}
There are many DNN mapper proposals utilizing different optimization methods, such as simulated annealing in TVM~\cite{autotvm}, RL in FlexTensor~\cite{flextensor} and ConfuciuX~\cite{confx}, Mixed Integer Programming in CoSA~\cite{cosa}, domain-specific GA in GAMMA~\cite{gamma}, gradient-based surrogate model in MindMapping~\cite{mindmapping}, and many others. However, 1) all the previously mentioned prior arts are mappers for intra-layer map-space while \system is a mapper for the less explored inter-layer map-space, and 2) they are search-based methods while \system is an inference-based method. 

\textbf{Layer-Fusion.} 
Some compiler works also study "Operator-Fusion" such as ~\cite{ivanov2020data, dnnfusion}, however, we want to clarify that "Operator-Fusion" usually fuses CONV or MatMul layer with element-wise operators such as ReLU but not several CONVs, FCs, or MatMul as we refer to by "Layer-Fusion". "Layer-Fusion" has more complexity and is therefore often discussed in the context of inter-layer dataflow or mapping. 
Layer fusion is an orthogonal map-space to intra-layer mapping where most DNN mapper focuses on.
Some recent works have looked into this map-space such as fused-layer CNN~\cite{alwani2016fused}, TGPA~\cite{wei2018tgpa}, and FLAT~\cite{flat} and demonstrate the huge potential performance improvement by leveraging this one additional mapping dimension in DNN mapping. However, the previously mentioned prior arts focus on proposing a hand-tuned layer fusion strategy for the considered use-cases. In contrast, \system is a mapper which can work on different use-cases automatically.


\textbf{Transformers for RL Problems.}
Adapting RLs as sequence modeling problems has not been notably successful until some recent works, such as Decision Transformer~\cite{decision_transformer}. Further applications showcase its vitality. Boustati et. al. \cite{boustati2021transfer} examined and discovered some significant potentials in transfer learning, and \cite{ortega2021shaking} by DeepMind developed the idea further to human-AI interaction. This work, to the best of our knowledge, is the first work using a transformer with RL formulation to tackle DNN mapping problem. 

\section{Conclusion}
We propose a mapper, \system, for Layer-Fusion map-space, featuring its ability to find a solution at inference time, much faster compared to the best existing search-based mappers with compatible performance. \system can also generalize to unseen conditions and do transfer learning on a new DNN workload. 


\bibliographystyle{ACM-Reference-Format}
\bibliography{main}


\begin{thebibliography}{34}


\ifx \showCODEN    \undefined \def \showCODEN     #1{\unskip}     \fi
\ifx \showDOI      \undefined \def \showDOI       #1{#1}\fi
\ifx \showISBNx    \undefined \def \showISBNx     #1{\unskip}     \fi
\ifx \showISBNxiii \undefined \def \showISBNxiii  #1{\unskip}     \fi
\ifx \showISSN     \undefined \def \showISSN      #1{\unskip}     \fi
\ifx \showLCCN     \undefined \def \showLCCN      #1{\unskip}     \fi
\ifx \shownote     \undefined \def \shownote      #1{#1}          \fi
\ifx \showarticletitle \undefined \def \showarticletitle #1{#1}   \fi
\ifx \showURL      \undefined \def \showURL       {\relax}        \fi
\providecommand\bibfield[2]{#2}
\providecommand\bibinfo[2]{#2}
\providecommand\natexlab[1]{#1}
\providecommand\showeprint[2][]{arXiv:#2}

\bibitem[{Alwani \em et al.}(2016)]%
        {alwani2016fused}
\bibfield{author}{\bibinfo{person}{M. {Alwani \em et al.}}}
  \bibinfo{year}{2016}\natexlab{}.
\newblock \showarticletitle{Fused-layer CNN accelerators}. In
  \bibinfo{booktitle}{\emph{MICRO}}. IEEE, \bibinfo{pages}{1--12}.
\newblock


\bibitem[Boustati et~al\mbox{.}(2021)]%
        {boustati2021transfer}
\bibfield{author}{\bibinfo{person}{Ayman Boustati} {et~al\mbox{.}}}
  \bibinfo{year}{2021}\natexlab{}.
\newblock \bibinfo{title}{Transfer learning with causal counterfactual
  reasoning in Decision Transformers}.
\newblock
\newblock
\showeprint[arxiv]{2110.14355}~[cs.LG]


\bibitem[Chen et~al\mbox{.}(2021)]%
        {decision_transformer}
\bibfield{author}{\bibinfo{person}{Lili Chen} {et~al\mbox{.}}}
  \bibinfo{year}{2021}\natexlab{}.
\newblock \showarticletitle{Decision transformer: Reinforcement learning via
  sequence modeling}.
\newblock \bibinfo{journal}{\emph{arXiv preprint arXiv:2106.01345}}
  (\bibinfo{year}{2021}).
\newblock


\bibitem[Chen et~al\mbox{.}(2018)]%
        {autotvm}
\bibfield{author}{\bibinfo{person}{Tianqi Chen} {et~al\mbox{.}}}
  \bibinfo{year}{2018}\natexlab{}.
\newblock \showarticletitle{Learning to optimize tensor programs}. In
  \bibinfo{booktitle}{\emph{NeurIPS}}. \bibinfo{pages}{3389--3400}.
\newblock


\bibitem[Chen et~al\mbox{.}(2016)]%
        {eyeriss_isca}
\bibfield{author}{\bibinfo{person}{Yu-Hsin Chen} {et~al\mbox{.}}}
  \bibinfo{year}{2016}\natexlab{}.
\newblock \showarticletitle{{Eyeriss: A spatial architecture for
  energy-efficient dataflow for convolutional neural networks}}. In
  \bibinfo{booktitle}{\emph{ISCA}}.
\newblock


\bibitem[Google(2020)]%
        {edge_tpu}
\bibfield{author}{\bibinfo{person}{Google}.} \bibinfo{year}{2020}\natexlab{}.
\newblock \bibinfo{title}{Edge TPU}.
\newblock
\newblock
\urldef\tempurl%
\url{https://cloud.google.com/edge-tpu}
\showURL{%
\tempurl}


\bibitem[Hansen(2006)]%
        {cma}
\bibfield{author}{\bibinfo{person}{Nikolaus Hansen}.}
  \bibinfo{year}{2006}\natexlab{}.
\newblock \showarticletitle{The CMA evolution strategy: a comparing review}.
\newblock In \bibinfo{booktitle}{\emph{Towards a new evolutionary
  computation}}. \bibinfo{publisher}{Springer}, \bibinfo{pages}{75--102}.
\newblock


\bibitem[He et~al\mbox{.}(2016)]%
        {Resnet}
\bibfield{author}{\bibinfo{person}{Kaiming He} {et~al\mbox{.}}}
  \bibinfo{year}{2016}\natexlab{}.
\newblock \showarticletitle{Deep residual learning for image recognition}. In
  \bibinfo{booktitle}{\emph{CVPR}}. \bibinfo{pages}{770--778}.
\newblock


\bibitem[Hegde et~al\mbox{.}(2021)]%
        {mindmapping}
\bibfield{author}{\bibinfo{person}{Kartik Hegde}, \bibinfo{person}{Po-An Tsai},
  \bibinfo{person}{Sitao Huang}, \bibinfo{person}{Vikas Chandra},
  \bibinfo{person}{Angshuman Parashar}, {and} \bibinfo{person}{Christopher~W
  Fletcher}.} \bibinfo{year}{2021}\natexlab{}.
\newblock \showarticletitle{Mind mappings: enabling efficient
  algorithm-accelerator mapping space search}. In
  \bibinfo{booktitle}{\emph{Proceedings of the 26th ACM International
  Conference on Architectural Support for Programming Languages and Operating
  Systems}}. \bibinfo{pages}{943--958}.
\newblock


\bibitem[Hellwig and Beyer(2016)]%
        {tbpsa}
\bibfield{author}{\bibinfo{person}{Michael Hellwig} {and}
  \bibinfo{person}{Hans-Georg Beyer}.} \bibinfo{year}{2016}\natexlab{}.
\newblock \showarticletitle{Evolution under strong noise: A self-adaptive
  evolution strategy can reach the lower performance bound-the pccmsa-es}. In
  \bibinfo{booktitle}{\emph{International Conference on Parallel Problem
  Solving from Nature}}. Springer, \bibinfo{pages}{26--36}.
\newblock


\bibitem[Huang et~al\mbox{.}(2021)]%
        {cosa}
\bibfield{author}{\bibinfo{person}{Qijing Huang} {et~al\mbox{.}}}
  \bibinfo{year}{2021}\natexlab{}.
\newblock \showarticletitle{CoSA: Scheduling by Constrained Optimization for
  Spatial Accelerators}.
\newblock \bibinfo{journal}{\emph{arXiv preprint arXiv:2105.01898}}
  (\bibinfo{year}{2021}).
\newblock


\bibitem[Ivanov et~al\mbox{.}(2020)]%
        {ivanov2020data}
\bibfield{author}{\bibinfo{person}{Andrei Ivanov} {et~al\mbox{.}}}
  \bibinfo{year}{2020}\natexlab{}.
\newblock \showarticletitle{Data Movement Is All You Need: A Case Study on
  Optimizing Transformers}.
\newblock \bibinfo{journal}{\emph{arXiv e-prints}} (\bibinfo{year}{2020}),
  \bibinfo{pages}{arXiv--2007}.
\newblock


\bibitem[Jouppi et~al\mbox{.}(2020)]%
        {tpu_spec}
\bibfield{author}{\bibinfo{person}{Norman~P. Jouppi}, \bibinfo{person}{Doe~Hyun
  Yoon}, \bibinfo{person}{George Kurian}, \bibinfo{person}{Sheng Li},
  \bibinfo{person}{Nishant Patil}, \bibinfo{person}{James Laudon},
  \bibinfo{person}{Cliff Young}, {and} \bibinfo{person}{David Patterson}.}
  \bibinfo{year}{2020}\natexlab{}.
\newblock \showarticletitle{A Domain-Specific Supercomputer for Training Deep
  Neural Networks}.
\newblock \bibinfo{journal}{\emph{Commun. ACM}} \bibinfo{volume}{63},
  \bibinfo{number}{7} (\bibinfo{date}{June} \bibinfo{year}{2020}),
  \bibinfo{pages}{67–78}.
\newblock
\showISSN{0001-0782}
\urldef\tempurl%
\url{https://doi.org/10.1145/3360307}
\showDOI{\tempurl}


\bibitem[Kao et~al\mbox{.}(2020a)]%
        {confx}
\bibfield{author}{\bibinfo{person}{Sheng-Chun Kao} {et~al\mbox{.}}}
  \bibinfo{year}{2020}\natexlab{a}.
\newblock \showarticletitle{ConfuciuX: Autonomous Hardware Resource Assignment
  for DNN Accelerators using Reinforcement Learning}. In
  \bibinfo{booktitle}{\emph{MICRO}}. IEEE, \bibinfo{pages}{1--14}.
\newblock


\bibitem[Kao et~al\mbox{.}(2020b)]%
        {gamma}
\bibfield{author}{\bibinfo{person}{Sheng-Chun Kao} {et~al\mbox{.}}}
  \bibinfo{year}{2020}\natexlab{b}.
\newblock \showarticletitle{GAMMA: Automating the HW mapping of DNN models on
  accelerators via genetic algorithm}. In \bibinfo{booktitle}{\emph{ICCAD}}.
  IEEE, \bibinfo{pages}{1--9}.
\newblock


\bibitem[Kao et~al\mbox{.}(2021)]%
        {flat}
\bibfield{author}{\bibinfo{person}{Sheng-Chun Kao} {et~al\mbox{.}}}
  \bibinfo{year}{2021}\natexlab{}.
\newblock \showarticletitle{An Optimized Dataflow for Mitigating Attention
  Performance Bottlenecks}.
\newblock \bibinfo{journal}{\emph{arXiv preprint arXiv:2107.06419}}
  (\bibinfo{year}{2021}).
\newblock


\bibitem[Kennedy and Eberhart(1995)]%
        {pso_paper}
\bibfield{author}{\bibinfo{person}{James Kennedy} {and}
  \bibinfo{person}{Russell Eberhart}.} \bibinfo{year}{1995}\natexlab{}.
\newblock \showarticletitle{Particle swarm optimization}. In
  \bibinfo{booktitle}{\emph{Proceedings of ICNN'95-International Conference on
  Neural Networks}}, Vol.~\bibinfo{volume}{4}. IEEE,
  \bibinfo{pages}{1942--1948}.
\newblock


\bibitem[Kwon et~al\mbox{.}(2018)]%
        {maestro}
\bibfield{author}{\bibinfo{person}{Hyoukjun Kwon}, \bibinfo{person}{Michael
  Pellauer}, {and} \bibinfo{person}{Tushar Krishna}.}
  \bibinfo{year}{2018}\natexlab{}.
\newblock \showarticletitle{An Analytic Model for Cost-Benefit Analysis of
  Dataflows in DNN Accelerators}.
\newblock \bibinfo{journal}{\emph{arXiv preprint arXiv:1805.02566}}
  (\bibinfo{year}{2018}).
\newblock


\bibitem[Lillicrap et~al\mbox{.}(2015)]%
        {ddpg}
\bibfield{author}{\bibinfo{person}{Timothy~P Lillicrap} {et~al\mbox{.}}}
  \bibinfo{year}{2015}\natexlab{}.
\newblock \showarticletitle{Continuous control with deep reinforcement
  learning}.
\newblock \bibinfo{journal}{\emph{arXiv preprint arXiv:1509.02971}}
  (\bibinfo{year}{2015}).
\newblock


\bibitem[Mnih et~al\mbox{.}(2016)]%
        {a2c}
\bibfield{author}{\bibinfo{person}{Volodymyr Mnih} {et~al\mbox{.}}}
  \bibinfo{year}{2016}\natexlab{}.
\newblock \showarticletitle{Asynchronous methods for deep reinforcement
  learning}. In \bibinfo{booktitle}{\emph{ICML}}. \bibinfo{pages}{1928--1937}.
\newblock


\bibitem[Niu et~al\mbox{.}(2021)]%
        {dnnfusion}
\bibfield{author}{\bibinfo{person}{Wei Niu} {et~al\mbox{.}}}
  \bibinfo{year}{2021}\natexlab{}.
\newblock \showarticletitle{DNNFusion: accelerating deep neural networks
  execution with advanced operator fusion}. In
  \bibinfo{booktitle}{\emph{PLDI'21}}. \bibinfo{pages}{883--898}.
\newblock


\bibitem[NVIDIA(2018)]%
        {nvdla}
\bibfield{author}{\bibinfo{person}{NVIDIA}.} \bibinfo{year}{2018}\natexlab{}.
\newblock \bibinfo{title}{{NVIDIA Deep Learning Accelerator (NVDLA)}}.
\newblock \bibinfo{howpublished}{\url{https://nvldla.org}}.
\newblock


\bibitem[Ortega et~al\mbox{.}(2021)]%
        {ortega2021shaking}
\bibfield{author}{\bibinfo{person}{Pedro~A. Ortega} {et~al\mbox{.}}}
  \bibinfo{year}{2021}\natexlab{}.
\newblock \bibinfo{title}{Shaking the foundations: delusions in sequence models
  for interaction and control}.
\newblock
\newblock
\showeprint[arxiv]{2110.10819}~[cs.LG]


\bibitem[Pelikan et~al\mbox{.}(1999)]%
        {bayes}
\bibfield{author}{\bibinfo{person}{Martin Pelikan} {et~al\mbox{.}}}
  \bibinfo{year}{1999}\natexlab{}.
\newblock \showarticletitle{BOA: The Bayesian optimization algorithm}. In
  \bibinfo{booktitle}{\emph{Proceedings of the genetic and evolutionary
  computation conference GECCO-99}}, Vol.~\bibinfo{volume}{1}.
  \bibinfo{pages}{525--532}.
\newblock


\bibitem[Price(2013)]%
        {de}
\bibfield{author}{\bibinfo{person}{Kenneth~V Price}.}
  \bibinfo{year}{2013}\natexlab{}.
\newblock \showarticletitle{Differential evolution}.
\newblock In \bibinfo{booktitle}{\emph{Handbook of Optimization}}.
  \bibinfo{publisher}{Springer}, \bibinfo{pages}{187--214}.
\newblock


\bibitem[Rapin and Teytaud(2018)]%
        {nevergrad}
\bibfield{author}{\bibinfo{person}{J. Rapin} {and} \bibinfo{person}{O.
  Teytaud}.} \bibinfo{year}{2018}\natexlab{}.
\newblock \bibinfo{title}{{Nevergrad - A gradient-free optimization platform}}.
\newblock
  \bibinfo{howpublished}{\url{https://GitHub.com/FacebookResearch/Nevergrad}}.
\newblock


\bibitem[Sandler et~al\mbox{.}(2018)]%
        {sandler2018mobilenetv2}
\bibfield{author}{\bibinfo{person}{Mark Sandler}, \bibinfo{person}{Andrew
  Howard}, \bibinfo{person}{Menglong Zhu}, \bibinfo{person}{Andrey Zhmoginov},
  {and} \bibinfo{person}{Liang-Chieh Chen}.} \bibinfo{year}{2018}\natexlab{}.
\newblock \showarticletitle{Mobilenetv2: Inverted residuals and linear
  bottlenecks}. In \bibinfo{booktitle}{\emph{Proceedings of the IEEE Conference
  on Computer Vision and Pattern Recognition}}. \bibinfo{pages}{4510--4520}.
\newblock


\bibitem[Simonyan and Zisserman(2014)]%
        {vgg}
\bibfield{author}{\bibinfo{person}{Karen Simonyan} {and}
  \bibinfo{person}{Andrew Zisserman}.} \bibinfo{year}{2014}\natexlab{}.
\newblock \showarticletitle{Very deep convolutional networks for large-scale
  image recognition}.
\newblock \bibinfo{journal}{\emph{arXiv preprint arXiv:1409.1556}}
  (\bibinfo{year}{2014}).
\newblock


\bibitem[Sutton et~al\mbox{.}(2000)]%
        {reinforce}
\bibfield{author}{\bibinfo{person}{Richard~S Sutton} {et~al\mbox{.}}}
  \bibinfo{year}{2000}\natexlab{}.
\newblock \showarticletitle{Policy gradient methods for reinforcement learning
  with function approximation}. In \bibinfo{booktitle}{\emph{Advances in neural
  information processing systems}}. \bibinfo{pages}{1057--1063}.
\newblock


\bibitem[Tan et~al\mbox{.}(2019)]%
        {mnasnet}
\bibfield{author}{\bibinfo{person}{Mingxing Tan}, \bibinfo{person}{Bo Chen},
  \bibinfo{person}{Ruoming Pang}, \bibinfo{person}{Vijay Vasudevan},
  \bibinfo{person}{Mark Sandler}, \bibinfo{person}{Andrew Howard}, {and}
  \bibinfo{person}{Quoc~V Le}.} \bibinfo{year}{2019}\natexlab{}.
\newblock \showarticletitle{Mnasnet: Platform-aware neural architecture search
  for mobile}. In \bibinfo{booktitle}{\emph{Proceedings of the IEEE Conference
  on Computer Vision and Pattern Recognition}}. \bibinfo{pages}{2820--2828}.
\newblock


\bibitem[Vaswani et~al\mbox{.}(2017)]%
        {vaswani2017attention}
\bibfield{author}{\bibinfo{person}{Ashish Vaswani} {et~al\mbox{.}}}
  \bibinfo{year}{2017}\natexlab{}.
\newblock \showarticletitle{Attention is all you need}. In
  \bibinfo{booktitle}{\emph{NeurIPS}}. \bibinfo{pages}{5998--6008}.
\newblock


\bibitem[Wei et~al\mbox{.}(2018)]%
        {wei2018tgpa}
\bibfield{author}{\bibinfo{person}{Xuechao Wei} {et~al\mbox{.}}}
  \bibinfo{year}{2018}\natexlab{}.
\newblock \showarticletitle{TGPA: tile-grained pipeline architecture for low
  latency CNN inference}. In \bibinfo{booktitle}{\emph{ICCAD}}.
  \bibinfo{pages}{1--8}.
\newblock


\bibitem[Xiao et~al\mbox{.}(2021)]%
        {hasco}
\bibfield{author}{\bibinfo{person}{Qingcheng Xiao} {et~al\mbox{.}}}
  \bibinfo{year}{2021}\natexlab{}.
\newblock \showarticletitle{HASCO: Towards Agile HArdware and Software
  CO-design for Tensor Computation}.
\newblock \bibinfo{journal}{\emph{arXiv preprint arXiv:2105.01585}}
  (\bibinfo{year}{2021}).
\newblock


\bibitem[Zheng et~al\mbox{.}(2020)]%
        {flextensor}
\bibfield{author}{\bibinfo{person}{Size Zheng} {et~al\mbox{.}}}
  \bibinfo{year}{2020}\natexlab{}.
\newblock \showarticletitle{Flextensor: An automatic schedule exploration and
  optimization framework for tensor computation on heterogeneous system}. In
  \bibinfo{booktitle}{\emph{ASPLOS'20}}. \bibinfo{pages}{859--873}.
\newblock


\end{thebibliography}

\end{document}